\documentclass[conference, a4paper]{IEEEtran}
\usepackage[utf8]{inputenc} 
\usepackage[T1]{fontenc}
\usepackage{url}              
\usepackage{cite}             

\usepackage[cmex10]{amsmath}  
\usepackage{shorthands}
\usepackage{mleftright}       
\mleftright                   

\usepackage{graphicx}         
\usepackage{booktabs}         
\usepackage{subcaption}

\usepackage{algorithm}
\usepackage{algorithmicx}
\usepackage{algpseudocode}

\usepackage{xcolor}

\begin{document}

\title{Federated Neural Compression Under Heterogeneous Data
} 


%
\author{%
  \IEEEauthorblockN{Eric Lei, Hamed Hassani, and Shirin Saeedi Bidokhti}
  \IEEEauthorblockA{Department of Electrical and Systems Engineering\\
                    University of Pennsylvania, Philadelphia, PA\\
                    {\{\texttt{elei}, \texttt{hassani}, \texttt{saeedi}\}\texttt{@seas.upenn.edu}}}
}

\maketitle

\begin{abstract}
  We discuss a federated learned compression problem, where the goal is to learn a compressor from real-world data which is scattered across clients and may be statistically heterogeneous, yet share a common underlying representation. We propose a distributed source model that encompasses both characteristics, and naturally suggests a compressor architecture that uses analysis and synthesis transforms shared by clients. Inspired by personalized federated learning methods, we employ an entropy model that is personalized to each client. This allows for a global latent space to be learned across clients, and personalized entropy models that adapt to the clients' latent distributions. We show empirically that this strategy outperforms  solely local methods, which indicates that learned compression also benefits from a shared global representation in statistically heterogeneous federated settings. 
  
\end{abstract}

\section{Introduction}
\label{sec:intro}

Traditional learned compression usually takes place in a centralized setting, where one compression model is learned from data collected from various sources and stored at a single location. The standard lossy neural compression models the centralized data as a \textit{single} source $\bx \sim P_X$ supported on $\mathcal{X}$. Learned compression, under this source assumption, can be set up through the lens of nonlinear transform coding (NTC) \cite{balle2020nonlinear}. NTC seeks to find an analysis transform $g_a : \mathcal{X} \rightarrow \mathcal{Y}$ that maps $\bx$ to a latent variable $\by$, and a synthesis transform $g_s : \mathcal{Y} \rightarrow \mathcal{X}$ that maps from the latent space back to the source/reconstruction space. The latent variable $\by$ is then quantized to $\hat{\by} = \nint{\by}$ such that each entry is quantized to the nearest integer, and entropy coded using likelihoods generated from some entropy model $\pyhat(\hat{\by})$. The goal is to minimize the operational rate-distortion trade-off
 \begin{equation}
\EE_{\bx} [-\log_2 \pyhat({\nint{g_a(\bx)}})] + \lambda\EE_{\bx}[\dist(\bx,g_s(\nint{g_a(\bx)})],
\label{eq:single_source}
\end{equation}
where $\lambda > 0$ controls the trade-off. The transform functions $g_a$, $g_s$, as well as the entropy model are all parameterized using neural networks. 


In practice, NTC models are typically trained on aggregated datasets such as MS-COCO \cite{lin2014microsoft} or ImageNet \cite{imagenet}, which are collected from many clients into one location and form the samples from the source $\bx$. Then, a single model is trained on this source using the objective in \eqref{eq:single_source}. This centralized approach has been successful in a wide variety of applications and modalities such as image compression \cite{balle2018variational, minnen2018joint, minnen2020channel}, compression of channel state information (CSI) in wireless communications \cite{yang2019deep, ravula2021}, and in audio compression \cite{defossez2022high}. 

However, in many cases, a centralized approach is not necessarily feasible or applicable to the end user's setting. Rather than already existing in a central location, the data may be scattered across clients and it may not be feasible to centrally collect it due to limitations such as privacy constraints. Moreover, it can be expensive to collect all data at a centralized location when there are many clients. These challenges motivate analyzing learned compression from a distributed setup where multiple clients all wish to learn a compressor for their respective data in a federated fashion, potentially with the help of a central server. 

One immediate challenge that arises is that there are now potentially $n$ \textit{different} sources, where $n$ is the number of clients. The single source NTC setup in \eqref{eq:single_source} is, however, formulated for a single source. This parallels a well-known challenge in federated learning (FL), which is a related problem where clients wish to jointly learn a classifier or regressor from their data, where the data across clients are statistically heterogeneous, yet may share some underlying structure. For example, medical images collected using different equipment or in different locations may be modeled by different source distributions; yet, they are all fundamentally the same type of image. In the context of data compression, a natural question arises: how to model common structure across statistically heterogeneous sources? Furthermore, how can this shared structure be leveraged by the clients in order to learn good lossy compressors for each client?

For the former question, one difference that precludes direct application of the FL setup is that in FL, the statistical heterogeneity is modeled within the labels of the data distribution, not the features (e.g., pixel values in images). Learned compression is, however, an unsupervised task, and has no labeled data. Thus, all statistical heterogeneity and common structure that may exist across client data needs to be modeled solely within the features. In our work, we take these considerations into account by modelling the clients' sources as distributions induced by a shared generative function $f$ applied to $n$ independent latent sources. This shared generative function $f$ ensures that all sources share a common feature space.

For the latter question, a naive approach to solve this problem is to try to learn a single NTC-based compressor, where clients send model updates to the server at each round, and the server computes the average of the received models. This approach would extend FedAvg \cite{mcmahan2017communication} in FL to this federated compression setting. However, under our source heterogeneity assumption, it is more natural to learn client-specific compressors that are better tuned to each client's source distribution. One client-specific solution is to train a local model for each client. A challenge this approach faces in practice is that the number of samples each client has is small relative to the number of clients. Under our data heterogeneity model, however, there is intrinsic structure in the feature space defined by the shared function $f$, which can be leveraged across client data. We thus propose a novel solution with NTC models: learn globally shared analysis and synthesis transforms, with client-specific entropy models. This approach allows the shared analysis and synthesis transforms to try and undo the shared function $f$, extracting the client-specific latent source, which is then compressed using client-specific entropy models.

Our contributions are as follows. 
\begin{enumerate}
    \item We propose a federated learned compression framework that simultaneously encompasses statistical heterogeneity and shared intrinsic structure across clients. 
    \item We propose a novel method to learn client-specific neural compressors that leverages the shared structure across clients.
    \item We demonstrate empirically that learning a shared analysis and synthesis transform followed by locally optimized entropy models performs better than solely local NTC models. 
\end{enumerate}

\section{Related Work}

\subsection{Neural Compression}
The use of neural networks to design lossy compressors was initiated by merging quantization with autoencoder architectures \cite{toderici2016, balle2017endtoend, theis2017lossy, agustsson2017soft}. These methods, under the umbrella of NTC \cite{balle2020nonlinear}, operate by mapping the source to a latent space and back by using an analysis and synthesis transform parameterized with neural networks. Uniform quantization and entropy coding (using an additional learned entropy model) is then performed in the latent space, which is generally much lower in dimensionality. More recent work improve upon the entropy modeling to remove additional redundancy in the latent space. The scale hyperprior \cite{balle2018variational} transmits side information to generate different entropy models for each sample, with \cite{minnen2018joint} and \cite{minnen2020channel} building on top of this via autoregression of the latent variable. In single-image compression, these methods have shown to outperform recent handcrafted codecs such as HEVC \cite{BPG} and VVC \cite{VTM}. 

These methods are typically trained on data that has been pre-collected and stored centrally. One model is typically trained and deployed. In contrast, our work considers the case where the data resides across distributed clients such that collecting them centrally is not possible, and designing architectures to learn good compressors in such a setting.

\subsection{Federated Learning}
There have been many recent works discussing federated learning (FL), especially under client heterogeneity \cite{collins2021exploiting, mitra2021linear, karimireddy2020scaffold, reisizadeh2022straggler}. Most of these works have been applied in classification or regression settings, where each client $i$ has its own training data $\bx, \by \sim P_i$ and wishes to learn a good predictor that generalizes well on their respective distribution $P_i$. One popular approach is FedAvg \cite{mcmahan2017communication}, where a single model is sent to clients to update the model, before the central server averages their updates. Another approach is FedRep \cite{collins2021exploiting}, where a global feature extractor is learned on centralized data, and local client heads are jointly trained which use the extracted features to predict on local client data. Both methods have been shown to recover the underlying shared representation across clients in simple regression settings \cite{collins2022fedavg}. 

One difference in our setup is that learned compression does not assume labeled data. The assumptions of personalized FL are that heterogeneity resides in the labels, and hence the natural architecture to arise is a shared feature extractor followed by a predictor that is personalized. In contrast, we propose a different notion of heterogeneity with shared structure that purely resides in the features. We argue that this naturally leads to an architecture with shared analysis and synthesis transforms followed by personalized entropy model.

\subsection{Distributed Neural Compression}
There have also been recent works in designing distributed neural compression schemes \cite{mashhadi2021, mital2022neural, mital2023neural, whang2021neural}. These works are inspired by the information-theoretic results on compression with side information \cite{slepianwolf, wynerziv}, which say that if one has side information on a source to be compressed, an encoder that does not observe side information can perform just as well as one that does (in both cases the decoder observes the side information). However, this is slightly different from the federated setting. In contrast, we are interested in learning $n$ point-to-point lossy compressors (i.e., one for each client), where each compressor does not assume any side information from the other clients to be available. Rather than exploiting side information (i.e., a correlated source), we want to exploit underlying structure of the $n$ sources, despite being statistically independent. 

\section{Problem Formulation}
\subsection{Single-Source Neural Compression}
\begin{figure}[t]
    \centering
    \includegraphics[width=0.7\linewidth]{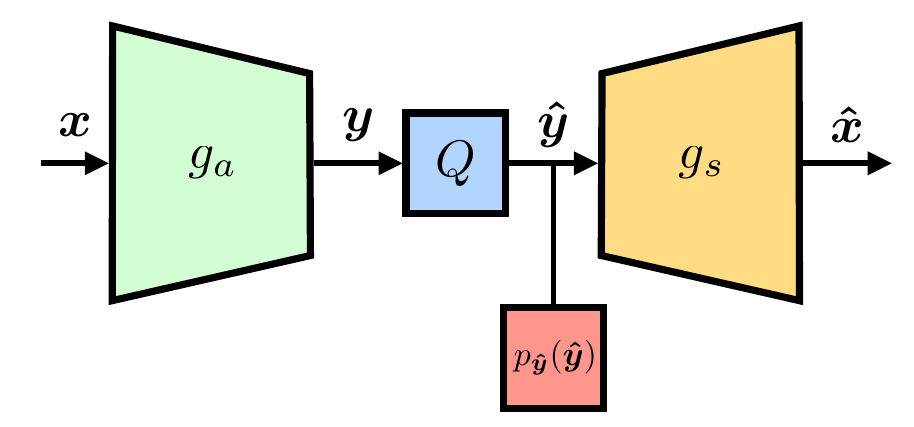}
    \caption{Single source neural compression.}
    \label{fig:NTC}
\end{figure}

As discussed in the introduction, classical neural compression models the data as a single source $\bx \sim P_X$, and attempts to learn a model that performs well in terms of rate and distortion defined w.r.t. $\bx$. The main intuition is that sources such as images have a low-dimensional latent space which can be extracted by the analysis and synthesis transforms. The low-dimensional latent representation is modelled probabilistically using a learned entropy model, which quantizes and entropy codes the latent variable. A figure describing this setup is shown in Fig.~\ref{fig:NTC}. 

This method encompasses centralized training, where datasets are aggregated ahead of time and a single model is trained using the combined data. 

\subsection{Federated Compression}

In the federated setting, instead of a single, effectively centralized source, we have $n$ sources $P_1,\dots,P_n$, with $\bx\expi \sim P_i$ consisting of the data seen by client $i$. Centralized training corresponds to aggregating all sources into the single source setup with $P_X = \pmix$. Typically, however, since the sources are heterogeneous across clients, it is better to have client-optimized models. Thus, the most general federated compression setup consists of $n$ separate compressors, such that each client $i$ has its own analysis and synthesis transforms $g_a\expi$, $g_s\expi$, as well as entropy model $\pyhat\expi$. Thus, the local federated objective consists of
\begin{align}
    \min_{\{g_a\expi, g_s\expi, \pyhat\expi\}_{i=1}^n} \frac{1}{n} \sum_{i=1}^n & \bigl ( \EE_{\bx\expi} [-\log_2 \pyhat\expi({\nint{g_a\expi(\bx\expi)}})] \nonumber \\
    +\lambda & \EE_{\bx\expi}[\dist(\bx\expi,g_s\expi(\nint{g_a\expi(\bx\expi)})] \bigr),
    \label{eq:local}
\end{align}
where the goal is minimize the rate and distortion averaged over the clients. We call the objective in \eqref{eq:local} Local-NTC, whose training algorithm can be found in Alg~\ref{alg:LocalOnlyNTC}.

\begin{algorithm}[t]
    \caption{Local-NTC}
    \begin{algorithmic}
        \Require Total iterations $T$, step size $\eta$, rate-distortion tradeoff $\lambda > 0$
        \State Initialize $\{g_a\expi\}_{i=1}^n, \{g_s\expi\}_{i=1}^n$, $\{\pyhat \expi\}_{i=1}^n$
        \For {$t=1,\dots,T$}
            \For {each client $i$}
                \State $(g_a \expi,g_s \expi, \pyhat \expi) \leftarrow (g_a \expi,g_s \expi, \pyhat \expi) - \eta \nabla_{(g_a \expi,g_s \expi, \pyhat \expi)}(R_n+\lambda D_n)$
            \EndFor
        \EndFor
    \end{algorithmic}
    \label{alg:LocalOnlyNTC}
\end{algorithm}

Rather than learn a local model for each client, results from federated learning suggest that heterogeneous data can potentially benefit from first learning a shared common representation. In the FedRep framework \cite{collins2021exploiting}, a common representation is extracted from a shared feature extractor, before individual client heads are learned to predict from this fixed representation. The intuition behind these works is the assumption that data heterogeneity exists primarily in the labels of the dataset, whereas the data samples (e.g., covariates or features) have a shared common structure. Learned compression, on the other hand, is a fully unsupervised task, and so both the heterogeneity and shared structure need to be modelled within the features.

\subsection{Heterogeneous Source Modelling}
\label{sec:heterogeneous_source}
As an example, if one wishes to compress similar types of images (e.g., medical images) that are collected using different equipment across the clients, one might expect that these images lie on the same image manifold despite the image distributions being different across clients. Mathematically, we can model $P_1,\dots,P_n$ as follows. Let $\{\bz\expi\}_{i=1}^n$ be the latent sources of randomness underlying the images sources, which could represents properties of the images such as class, orientation, lightning, or style. Then, the sources in the ambient space which the clients observe are generated as $\bx \expi = f(\bz\expi)$, for some fixed but unknown function $f$, which models the shared source manifold, but the statistically heterogeneity is modelled implicitly via the $\bz^{(i)}$'s.


In Local-NTC, the analysis and synthesis transforms at each client attempt to ``undo'' this generative function $f$ in order to recover $\bz\expi$ and model its probability density. However, each client may be data limited, and thus learning $f$ at every client's model may be difficult to accomplish. Instead, in order for each client $i$ to recover the underlying source, it may be easier to globally learn the function $f$ across clients. The NTC framework itself naturally possesses an architecture that can naturally support this heterogeneous source model, where the images across clients share an underlying latent representation.

\subsection{Sharing Analysis and Synthesis Transforms}

\begin{figure}[t]
    \centering
    \includegraphics[width=0.85\linewidth]{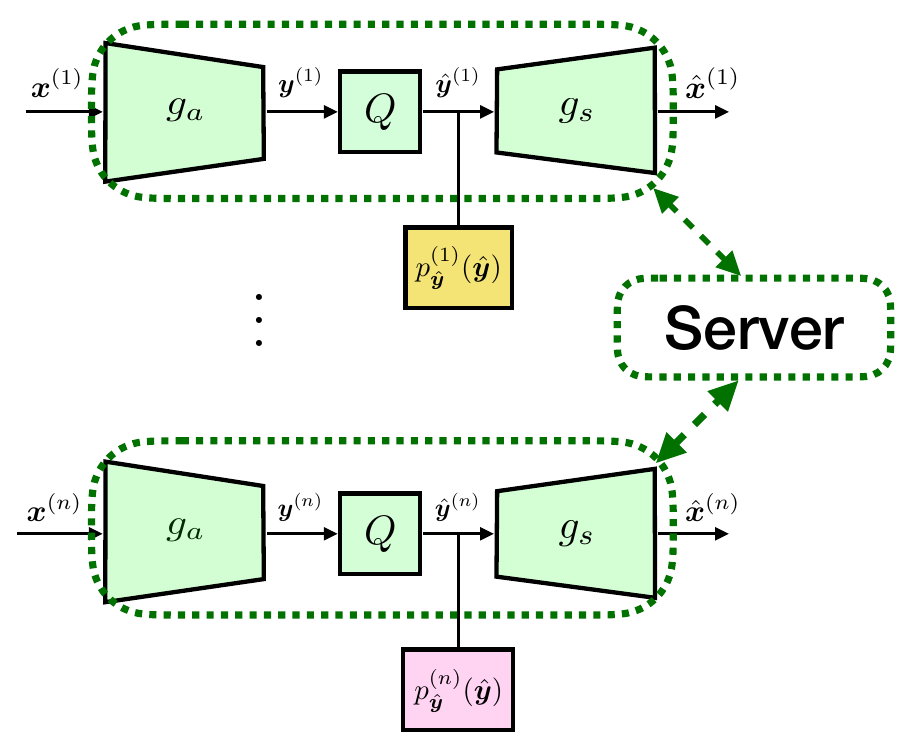}
    \caption{Federated neural compression under Fed-NTC framework. Analysis and synthesis transforms are learned globally through a central server, whereas the entropy models are learned per-client.}
    \label{fig:FedRepNTC}
\end{figure}

The analysis transform can be seen as a feature extractor that extracts a latent representation (typically of lower dimensionality), which is then quantized and entropy encoded according to a learned entropy model. One can thus view the ``prediction head'' of learned compression as the entropy coding part of the model. At the decoder side, the decompressed latent variable is transformed back to the reconstruction using the synthesis transform. Thus, if one learns a shared analysis and synthesis transform $g_a, g_s$ across clients, the induced distribution of the latent variables for client $i$ will be $g_a(\bx\expi)$. A client-specific entropy model $\pyhat\expi$ is then fine-tuned individually to more accurately model the distribution induced by $g_a(\bx\expi)$. Similar to federated learning, we expect this scheme to perform better than solely local training when the data per-client is small; this helps the analysis and synthesis transforms learn better feature extractors by leveraging all the data across clients. We call this setting Fed-NTC, where the objective is
\begin{align}
    \min_{g_a, g_s, \{\pyhat\expi\}_{i=1}^n} \frac{1}{n} \sum_{i=1}^n & \bigl ( \EE_{\bx\expi} [-\log_2 \pyhat\expi({\nint{g_a(\bx\expi)}})] \nonumber \\
    +\lambda & \EE_{\bx\expi}[\dist(\bx\expi,g_s(\nint{g_a(\bx\expi)})] \bigr),
    \label{eq:FedRepNTC}
\end{align}
which jointly optimizes the client-averaged rate and distortion $R_n + \lambda D_n$, where $R_n$ and $D_n$ are defined as
\begin{equation}
    R_n := \frac{1}{n} \sum_{i=1}^n  \EE_{\bx\expi} [-\log_2 \pyhat\expi({\nint{g_a(\bx\expi)}})],
\end{equation}
\begin{equation}
    D_n := \frac{1}{n} \sum_{i=1}^n  \EE_{\bx\expi}[\dist(\bx\expi,g_s(\nint{g_a(\bx\expi)})].
\end{equation}
Fig.~\ref{fig:FedRepNTC} describes the training framework. To optimize \eqref{eq:FedRepNTC}, one can alternate global updates, which are coordinated through the use of a server as shown in Fig.~\ref{fig:FedRepNTC}, with local entropy model updates. The full algorithm is shown in Alg.~\ref{alg:FedRepNTC}. 

\begin{algorithm}[t]
    \caption{Fed-NTC}
    \begin{algorithmic}
        \Require Participation rate $r$, communication rounds $T$, entropy model updates $T_p$, transform updates $T_g$, step size $\eta$, rate-distortion tradeoff $\lambda > 0$
        \State Initialize $g_a, g_s$, $\{\pyhat \expi\}_{i=1}^n$
        \For {$t=1,\dots,T$}
            \State Server sends $g_a, g_s$ to random fraction $r$ of clients $I_r$
            \For{$T_p$ iterations}
                \State $\pyhat \expi \leftarrow \pyhat \expi - \eta \nabla_{\pyhat \expi}(R_n+\lambda D_n), \quad \forall i\in I_r$
            \EndFor
            \State Initialize $(g_a \expi, g_s \expi) \leftarrow (g_a, g_s), \forall i \in I_r$
            \For{$T_g$ iterations}
                \State $(g_a \expi,g_s \expi) \leftarrow (g_a \expi,g_s \expi) - \eta \nabla_{(g_a \expi,g_s \expi)}(R_n+\lambda D_n)$
            \EndFor
            \State Each client $i$ sends $(g_a \expi,g_s \expi)$ back to server
            \State $(g_a, g_s) \leftarrow \frac{1}{rn}\sum_{i \in I_r} (g_a \expi,g_s \expi)$
        \EndFor
    \end{algorithmic}
    \label{alg:FedRepNTC}
\end{algorithm}

\subsection{Rate-Distortion Function For Heterogeneous Sources}
In this section, we explain why sharing analysis and synthesis transforms is principled, from a rate-distortion point of view. In general, the rate-distortion function of $n$ independent (but not necessarily identical) sources $\bx^{(i)} \sim P_i$ is achievable using a locally optimal code for each source, with an appropriate rate allocation across the $n$ sources. As such, the optimal trade-off between client-averaged rate and distortion $R^{\mathsf{fed}}(D)$ is given by \cite[Sec.~10.3.3]{CoverThomas}
\begin{equation}
    R^{\mathsf{fed}}(D) = \min_{D_1,\dots,D_n: \frac{1}{n}\sum_{i=1}^n D_i \leq D} \frac{1}{n} \sum_{i=1}^n R_i(D_i),
\end{equation}
where $R_i(D)$ is the rate-distortion function of $P_i$. This result appears to suggest that to obtain the best rate-distortion trade-off, it suffices to should learn $n$ separate compressors, one for each client, which is what Local-NTC does. However, under our proposed heterogeneous source model described in Sec.~\ref{sec:heterogeneous_source}, codes that share a global transform can also achieve $R^{\mathsf{fed}}(D)$ using the following structure. Setting $g_a = f^{-1}$ and $g_s = f$, first transform $\bx^{(i)}$ to $\bz^{(i)}$, which is compressed using an optimal code with respect to distortion function 
\begin{equation}
    \dist_z(\bz^{(i)}, \hat{\bz}^{(i)}) := \dist(f(\bz^{(i)}), f(\hat{\bz}^{(i)})).
\end{equation}
The compressed version of $\bz^{(i)}$ is transformed back using $f$. Thus, assuming that $\bz^{(i)}$ can be optimally compressed, both Local-NTC and Fed-NTC possess the architectures to perform optimal compression using this above scheme. However, in Local-NTC, each user's pair of transforms will need to learn $f^{-1}$ and $f$ individually, whereas in Fed-NTC, $f$ only needs to be learned for a single set of transforms at the global level, which should benefit learning-based compressors, especially when samples are limited. An algorithmic analysis of Fed-NTC is left for future work.

\section{Experiments}
We first discuss how we set up the experiments in order to introduce heterogeneity in the data, followed by their results. 
\label{sec:experiments}
\subsection{Experimental Setup}
\subsubsection{Datasets} To experimentally test the Fed-NTC framework, we test a federated setup on image compression of SVHN, and CIFAR10 datasets, which are all $32 \times 32$ RGB image datasets. While generally used for classification purposes, we use their class identities to introduce data heterogeneity across the clients. SVHN and CIFAR10 are both 10-class datasets. We assign $S$ classes to each client, on average, by using the non-i.i.d.\@ data partitioning method detailed in \cite{mcmahan2017communication}. We then vary $S$ in order to vary the heterogeneity across clients, where $S \in \{2, 5\}$. The number of training samples per user is fixed at $N/n$, where $N$ is the total number of training samples. This way of introducing heterogeneity can be viewed as the latent source $\bz\expi$ representing features of the subset of classes; for example, different poses and styles of the class object within the image frame. The generative mapping $f$ then projects these features to the image manifold. 

\subsubsection{Compression Models}
For the compression models, we use the NTC methods detailed in \cite{balle2020nonlinear}, with a factorized prior for $\pyhat$, which models each entry of $\pyhat$ independently using a single-variable density model parameterized by a neural network. In practice, all spatial elements along each channel dimension are modeled by the same density model in order to maintain translation-invariance across the entire model \cite{balle2018variational}. We add uniform noise to the $g_a(\bx)$ to serve as a proxy during training, with hard quantization and entropy coding during evaluation of the models. 

\subsubsection{Federated Setup}
For the federated setup, we use a total of $n=100$ clients in all image compression experiments. For the baseline model (Local-NTC), which consists of each client having its own model, each client trains its model on its own local data, e.g. in Alg.~\ref{alg:LocalOnlyNTC}, with no communication among the clients. For the proposed Fed-NTC method, we use a participation rate of $r=0.1$ in Alg.~\ref{alg:FedRepNTC} and a total of 100 rounds of communication. For evaluation, the number of bits needed to compress the hard-quantized latents, and the distortion are measured and averaged for the final 10 rounds of communications. 

\subsection{Results}

\begin{figure}[t]
     \centering
     \begin{subfigure}[b]{\linewidth}
         \centering
         \includegraphics[width=0.6\linewidth]{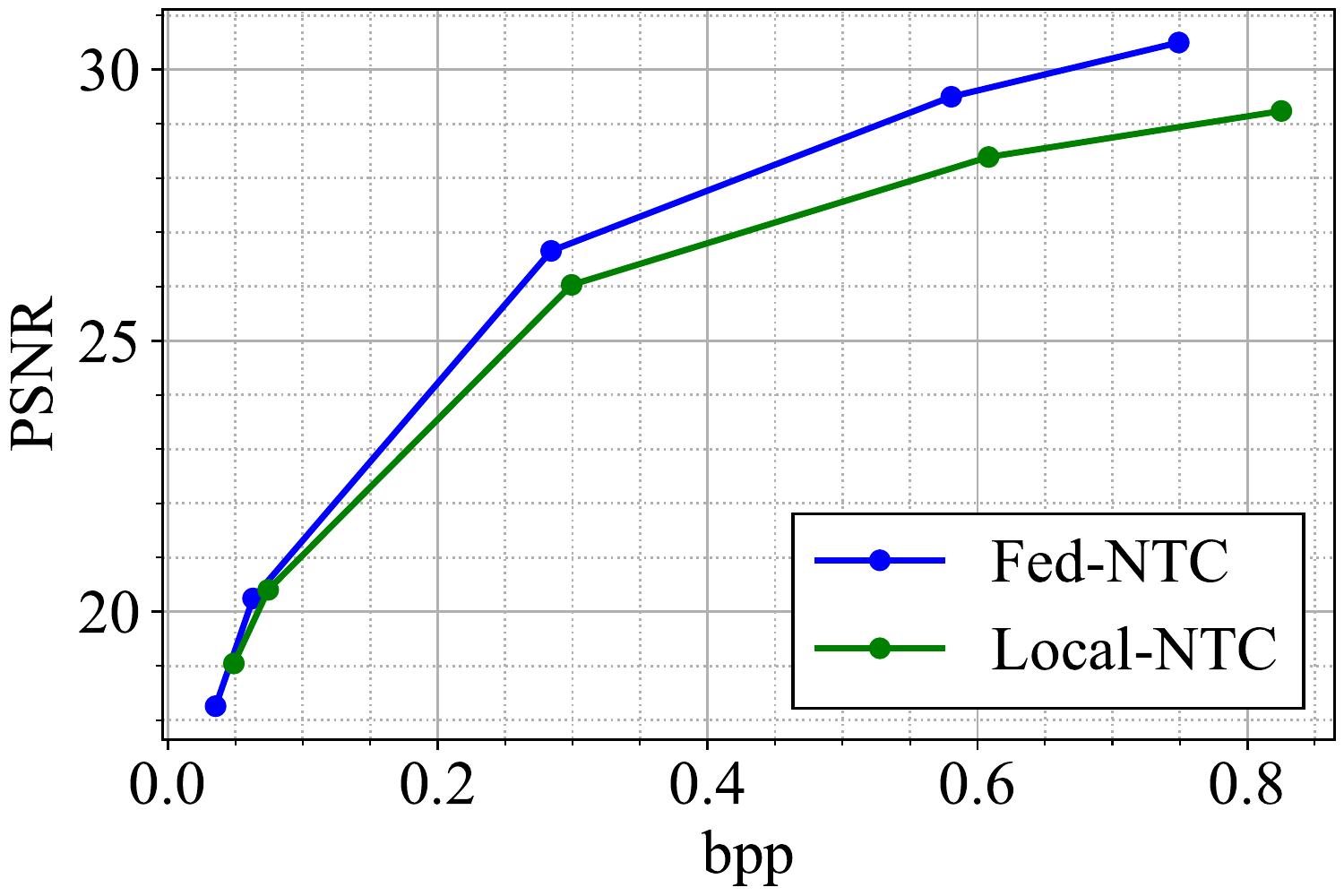}
         \caption{2 classes per client.}
         \label{fig:svhn_2}
     \end{subfigure}
     \vfill
     \begin{subfigure}[b]{\linewidth}
         \centering
         \includegraphics[width=0.59\linewidth]{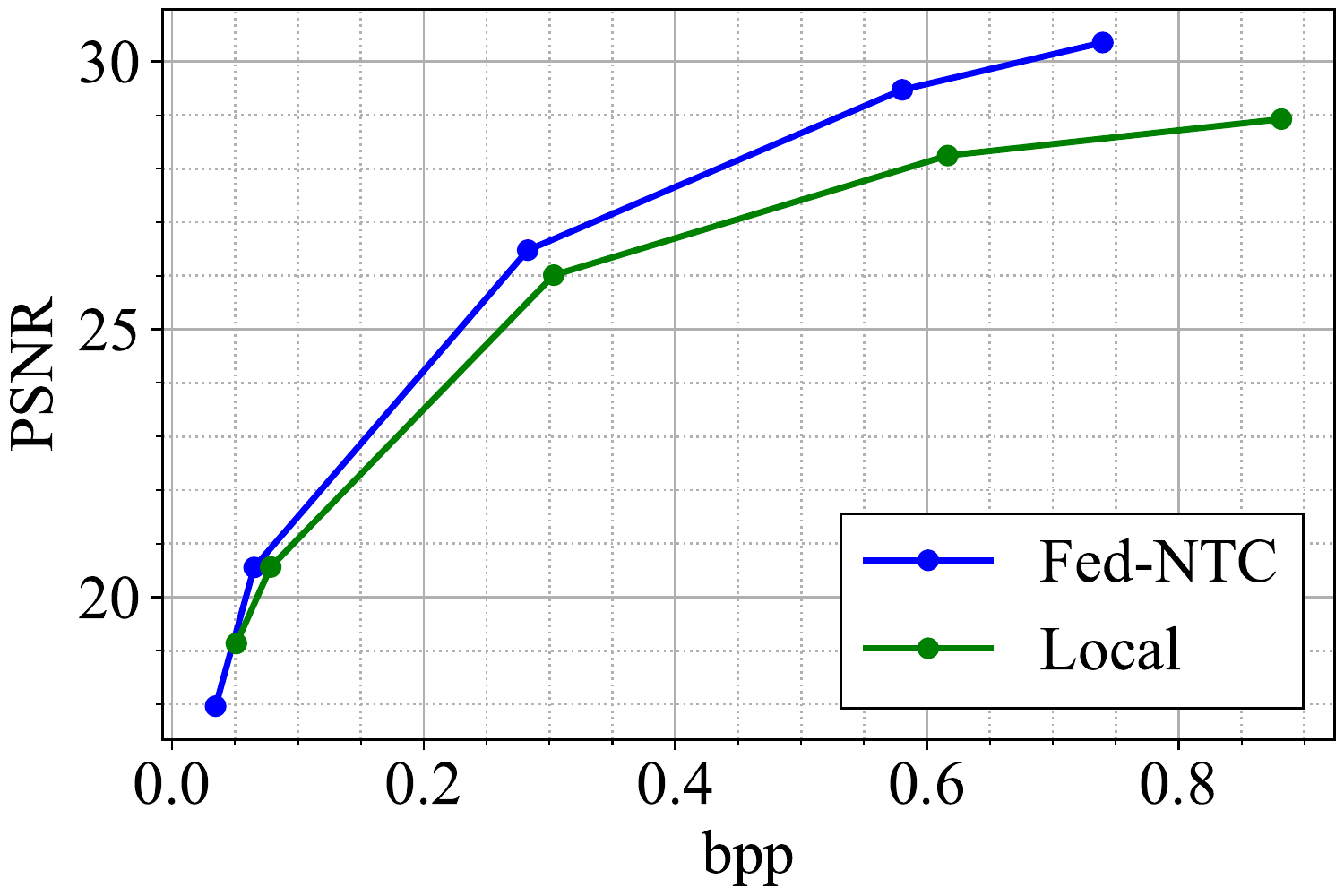}
         \caption{5 classes per client.}
         \label{fig:svhn_5}
     \end{subfigure}
        \caption{SVHN with 100 clients.}
        \label{fig:SVHN}
\end{figure}

\begin{figure}[t]
     \centering
     \begin{subfigure}[b]{\linewidth}
         \centering
         \includegraphics[width=0.6\linewidth]{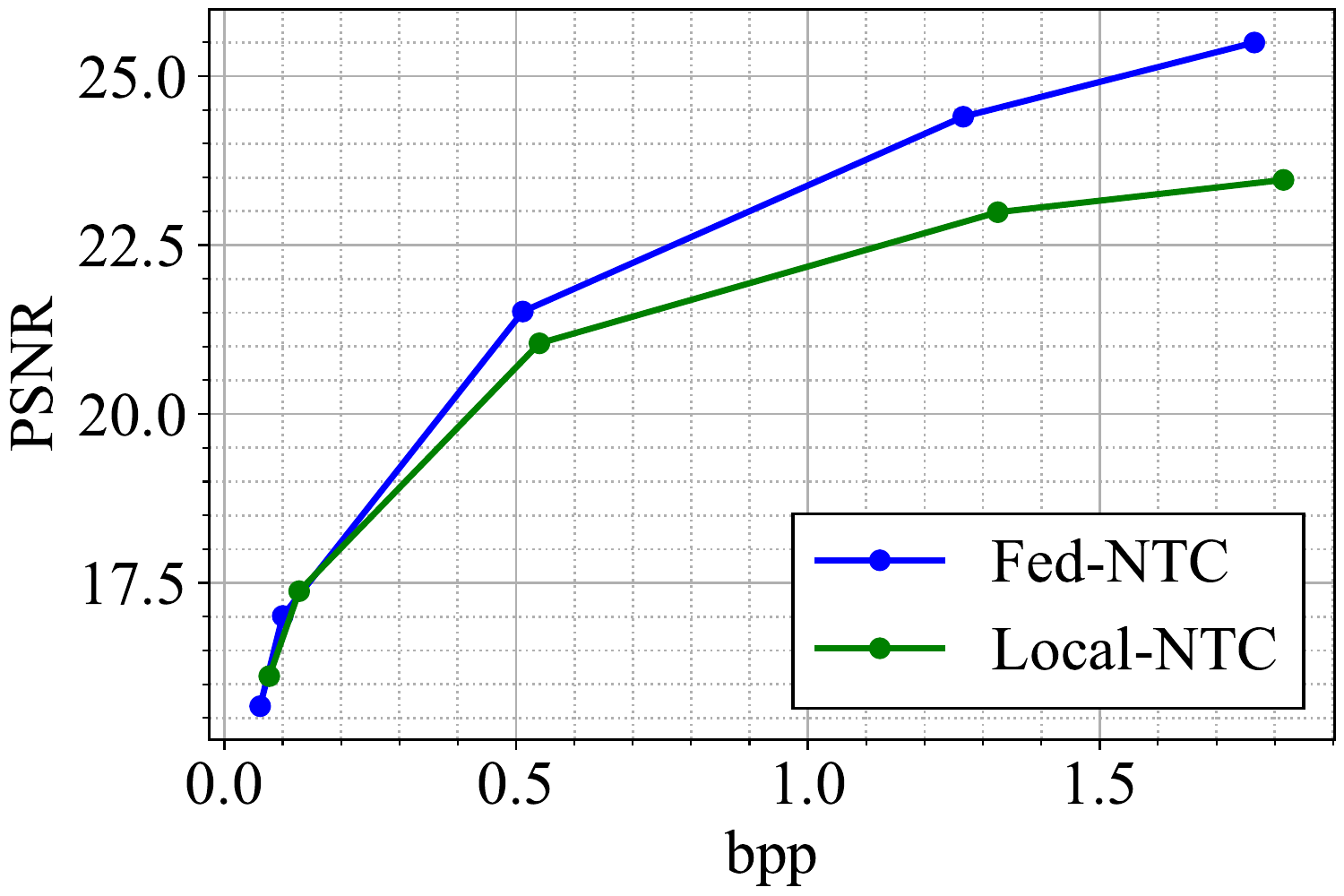}
         \caption{2 classes per client.}
         \label{fig:cifar10_2}
     \end{subfigure}
     \vfill
     \begin{subfigure}[b]{\linewidth}
         \centering
         \includegraphics[width=0.6\linewidth]{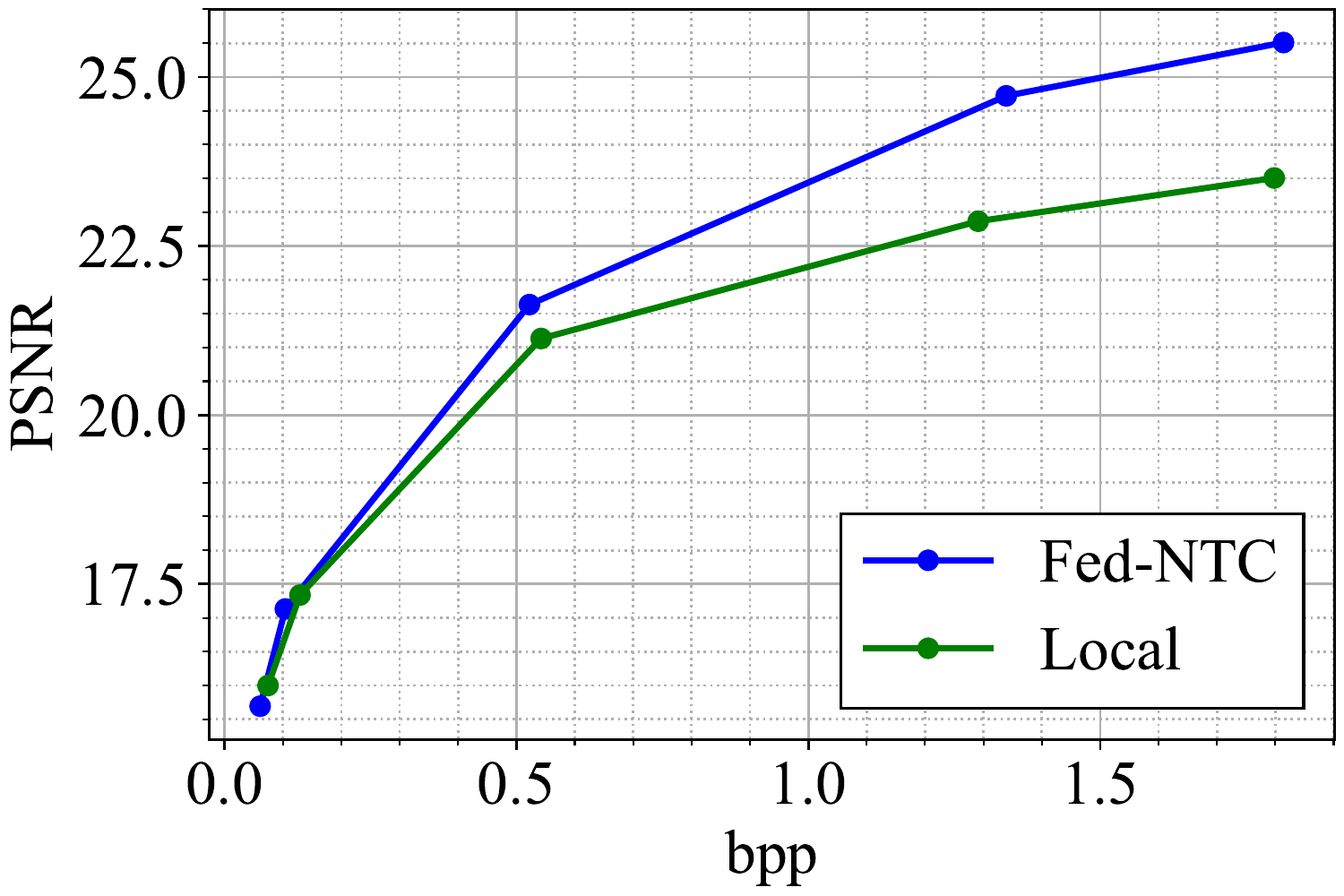}
         \caption{5 classes per client.}
         \label{fig:cifar10_5}
     \end{subfigure}
        \caption{CIFAR10 with 100 clients.}
        \label{fig:CIFAR10}
\end{figure}

\begin{figure}[t]
     \centering
     \begin{subfigure}[b]{\linewidth}
         \centering
         \includegraphics[width=0.57\linewidth]{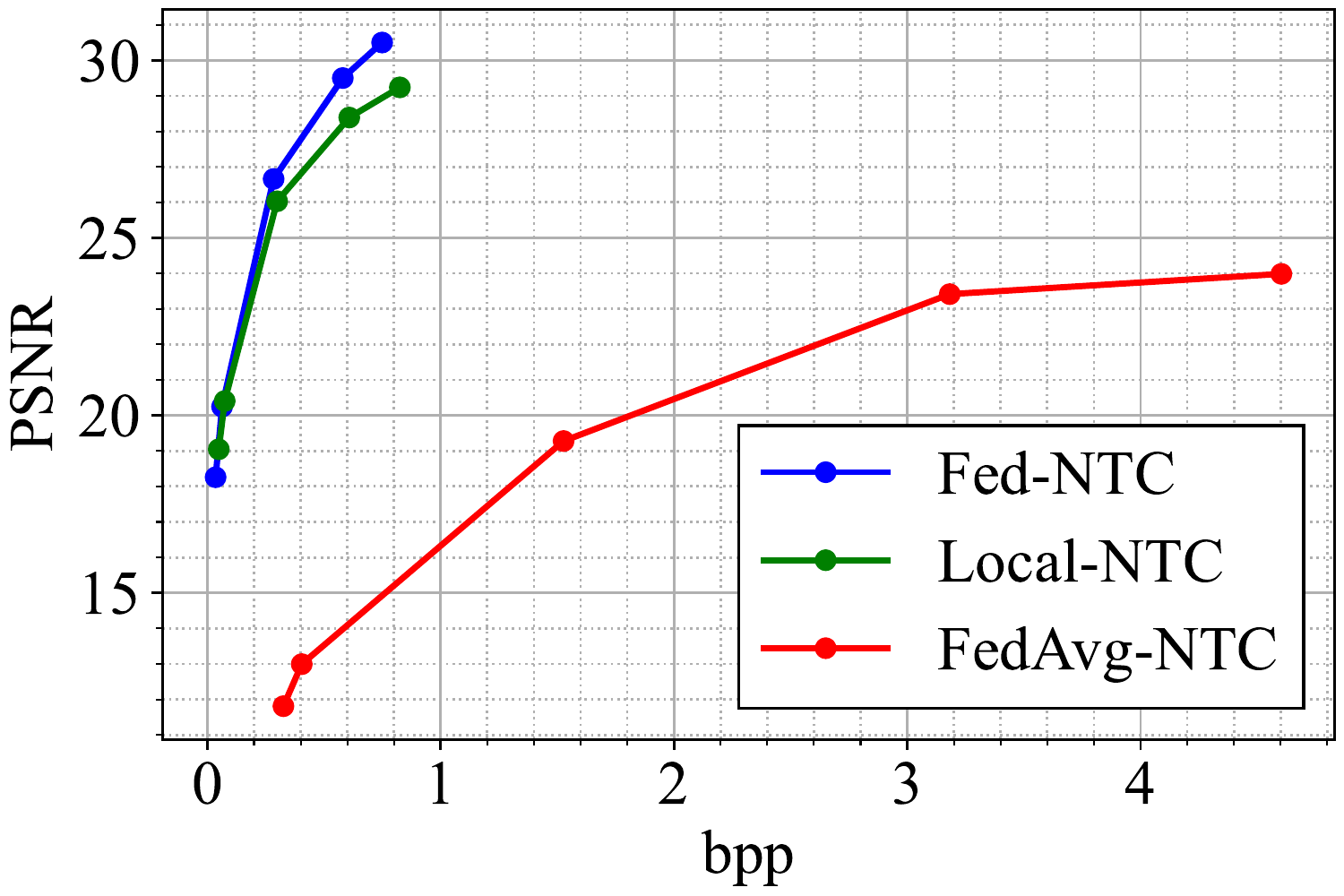}
         \caption{2 classes per client.}
     \end{subfigure}
     \vfill
     \begin{subfigure}[b]{\linewidth}
         \centering
         \includegraphics[width=0.57\linewidth]{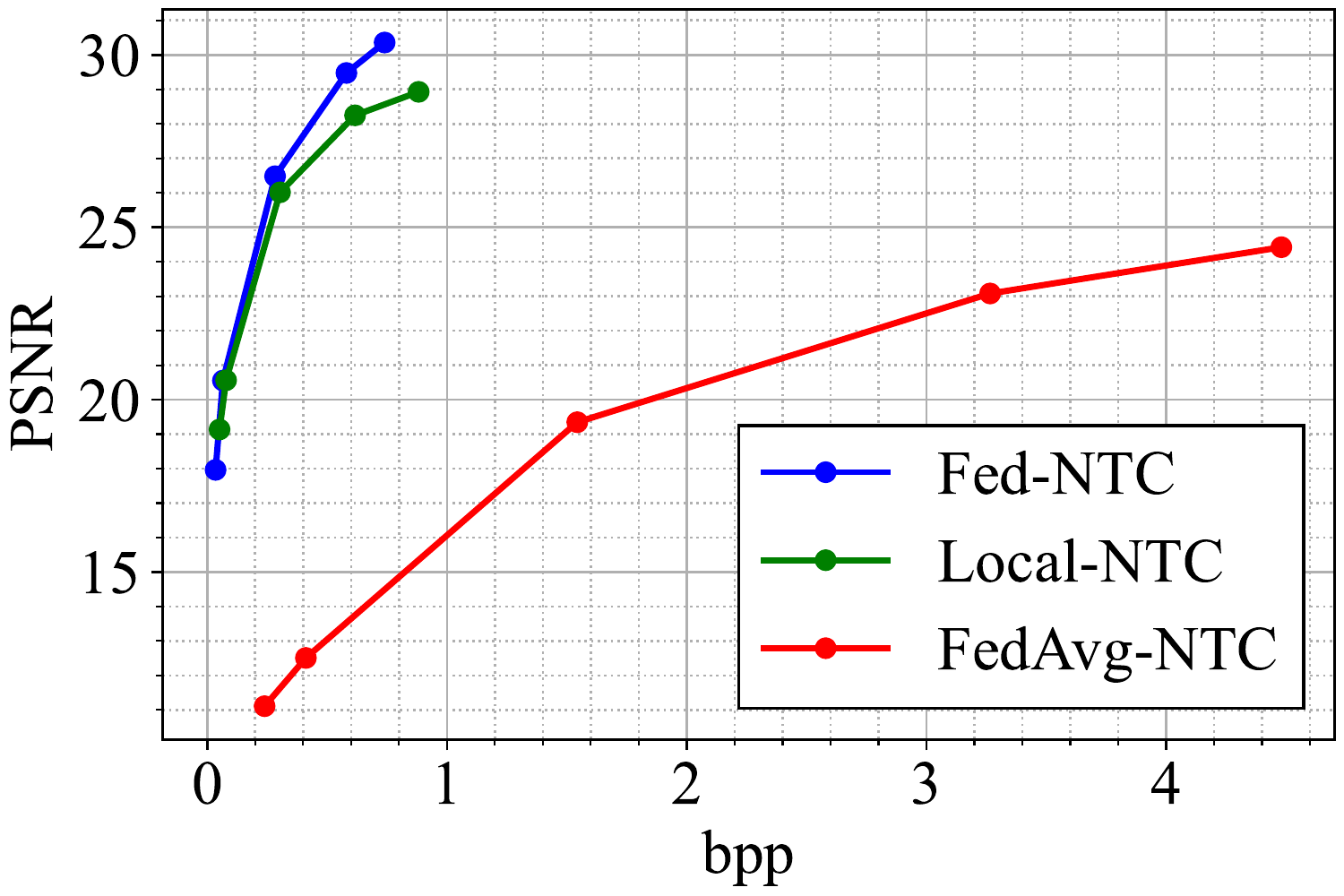}
         \caption{5 classes per client.}
     \end{subfigure}
        \caption{SVHN with 100 clients, compared with a single global NTC model trained under FedAvg.}
        \label{fig:SVHN_withfedavg}
\end{figure}

\begin{figure}[t]
     \centering
     \begin{subfigure}[b]{\linewidth}
         \centering
         \includegraphics[width=0.57\linewidth]{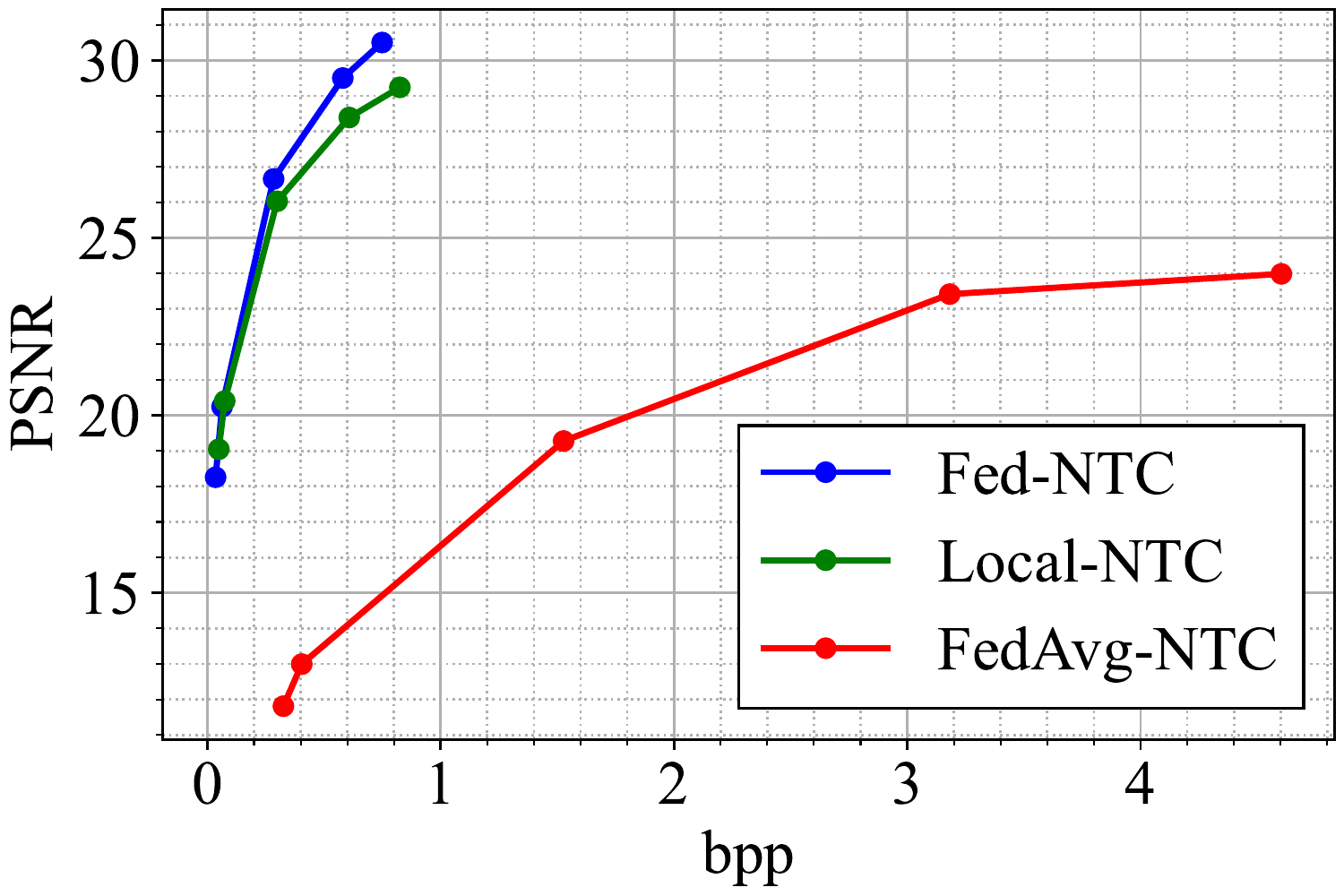}
         \caption{2 classes per client.}
     \end{subfigure}
     \vfill
     \begin{subfigure}[b]{\linewidth}
         \centering
         \includegraphics[width=0.57\linewidth]{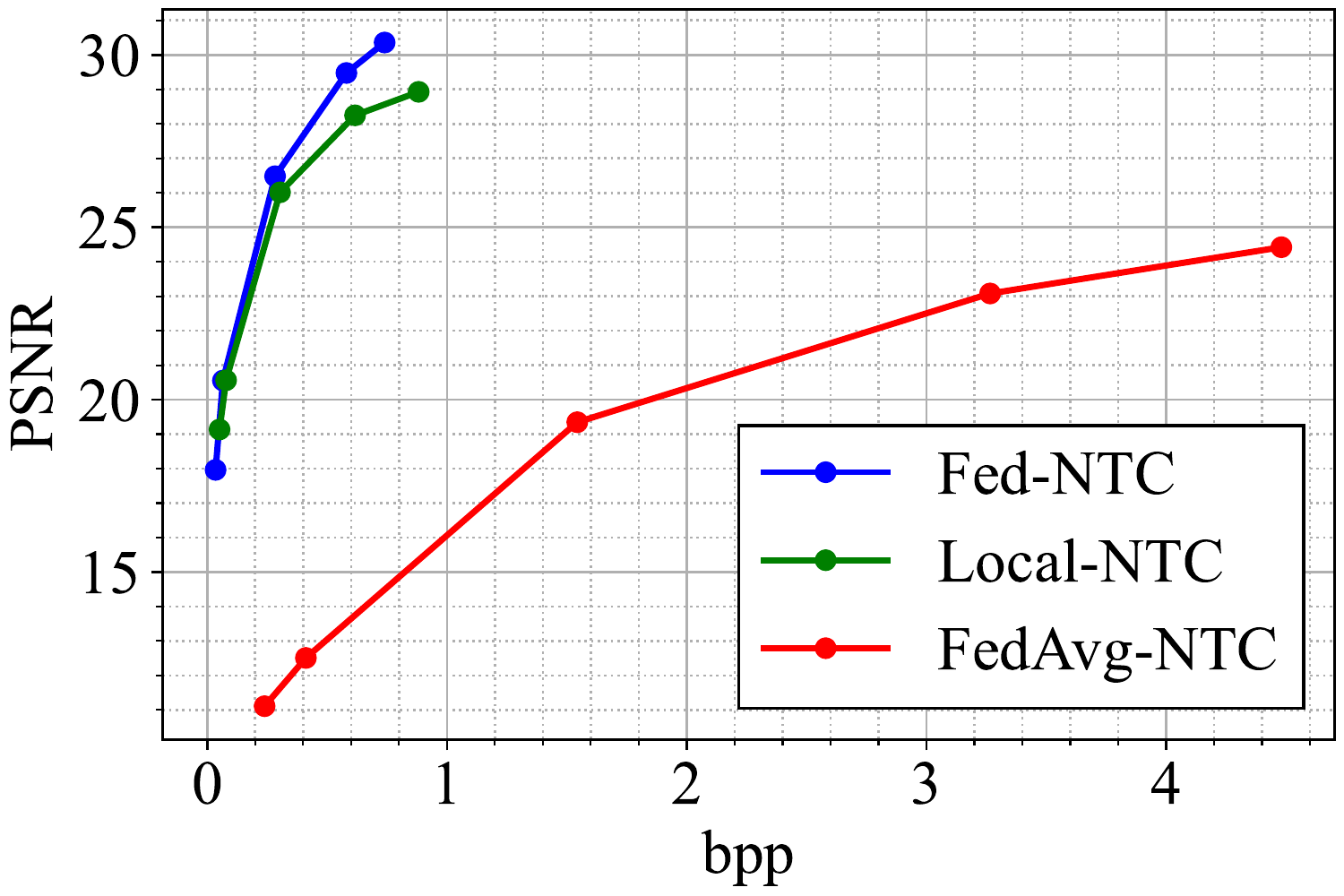}
         \caption{5 classes per client.}
     \end{subfigure}
        \caption{CIFAR10 with 100 clients, compared with a single global NTC model trained under FedAvg.}
        \label{fig:CIFAR10_withfedavg}
\end{figure}


In Figs.~\ref{fig:SVHN} and \ref{fig:CIFAR10}, we plot the comparisons of Fed-NTC with Local-NTC for CIFAR10 and SVHN, respectively. We observe that in almost all settings, Fed-NTC outperforms Local-NTC models. This indicates that learning a shared representation is able to compress the data across clients better at a variety of different levels of heterogeneity, and that shared generative function is a plausible model for the heterogeneous data with shared structure. The Fed-NTC models are able to recover the global functon $f$ by leveraging the data across clients. 

We additionally compare Fed-NTC with a model that shared a single global model across all users. This model, trained under the FedAvg scheme \cite{mcmahan2017communication}, alternates between 10 local updates for each client, before sending back to the server which averages all models. As shown in Figs.~\ref{fig:SVHN_withfedavg} and \ref{fig:CIFAR10_withfedavg}, such a model performs significantly worse than Fed-NTC and even local training, indicating that a single global model struggles with client heterogeneity in learned compression. We noted no significant difference if the number of local updates was reduced in an effort to reduce client drift. 

\begin{figure}[t]
     \centering
     \begin{subfigure}[b]{\linewidth}
         \centering
         \includegraphics[width=0.64\linewidth]{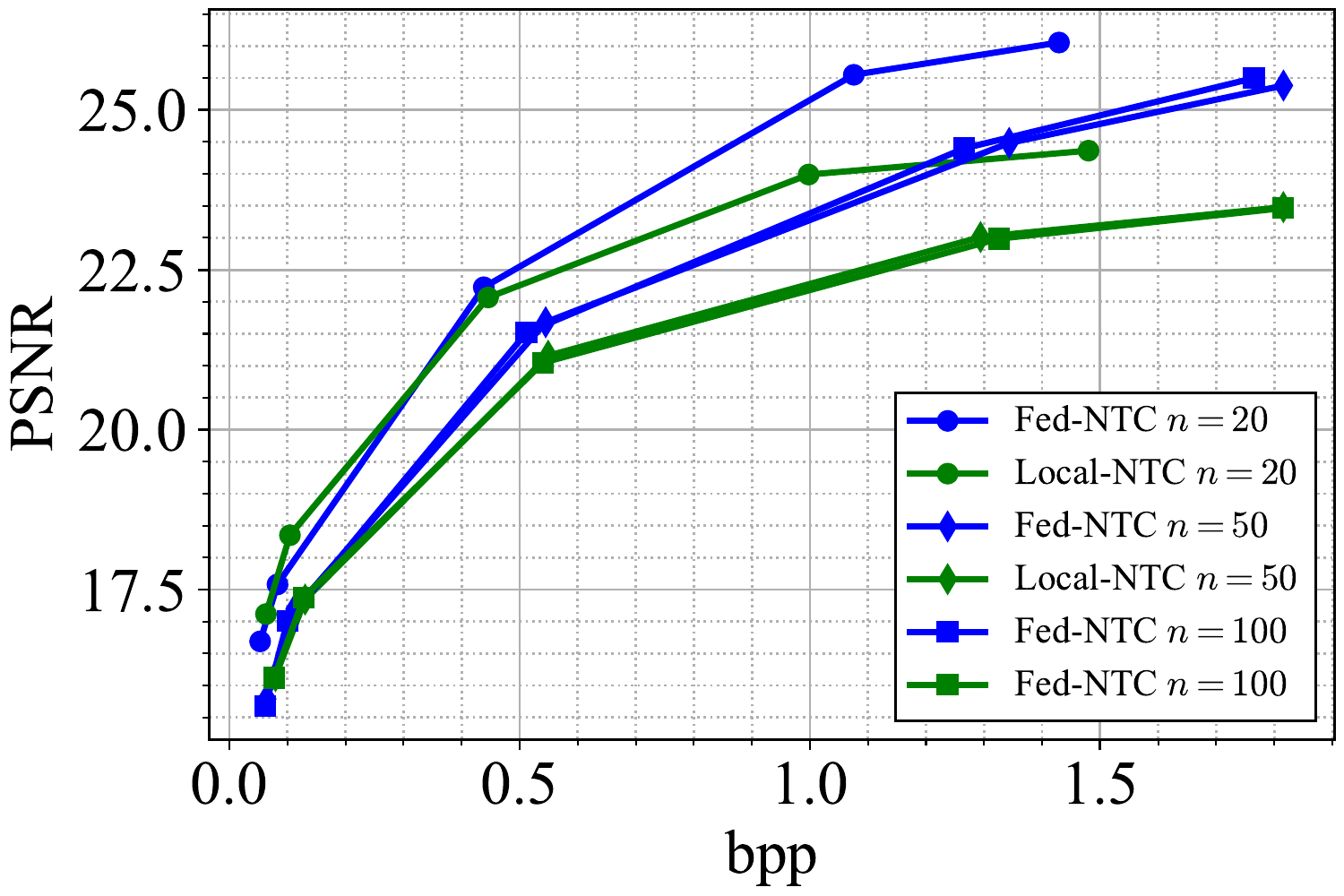}
         \caption{2 classes per client.}
     \end{subfigure}
     \vfill
     \begin{subfigure}[b]{\linewidth}
         \centering
         \includegraphics[width=0.64\linewidth]{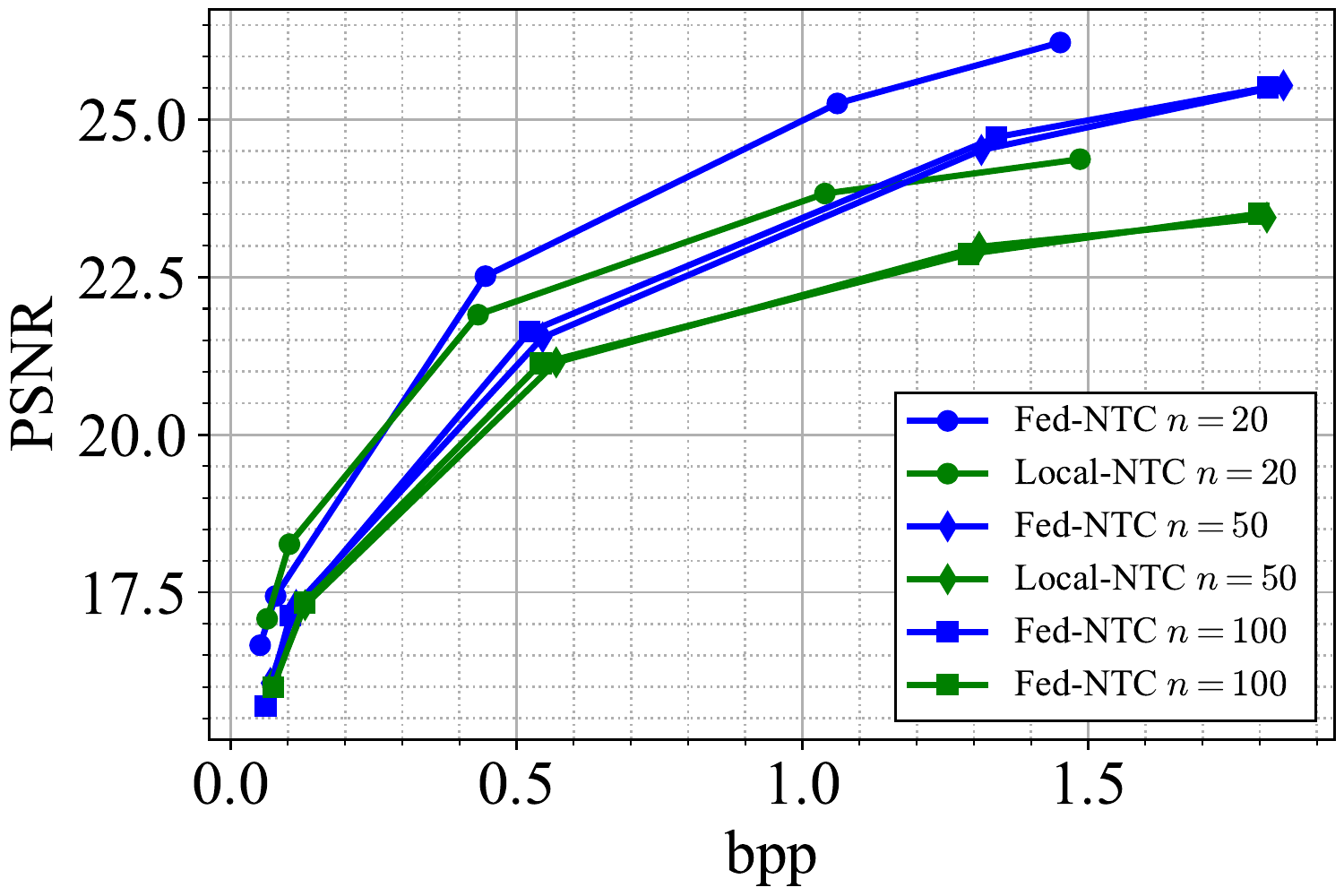}
         \caption{5 classes per client.}
     \end{subfigure}
        \caption{CIFAR10, varying number of users.}
        \label{fig:CIFAR10_varyusers}
\end{figure}

Finally, we run an additional experiment which varies the number of users $n$ (and thus the number of samples allotted to each user). In previous figures, we set $n=100$. In Fig.~\ref{fig:CIFAR10_varyusers}, we reduce the number of users to $n=50$ and $n=20$. We observe that at $n=50$ there is no significant change compared to $n=100$, but at $n=20$ both Fed-NTC and Local-NTC improve in terms of rate-distortion performance. 

\section{Conclusion}
\label{sec:conclusion}
We propose a federated framework for learned compression, where the data across clients may be heterogeneous. We propose a model for such data that has shared structure in the form of a common generative function. This model suggests that in compression, one should first extract a common latent space before client-specific entropy modelling takes place. Experimental results confirm that this assumption is true across a broad class of settings and this scheme is superior to solely local models. Potential avenues for future work include analyzing privacy aspects and algorithmic analysis of different federated schemes.                                                                                                                                                      


\bibliographystyle{IEEEtran}
\bibliography{ref}

\section*{Acknowledgments}
The work of Eric Lei was supported by a NSF Graduate Research Fellowship. The work of Shirin Saeedi Bidokhti was supported by NSF award 1910594 and an NSF CAREER award 2047482. The work of Hamed Hassani was supported by NSF award CIF-1943064. This work was also partially supported by the AI Institute for Learning-Enabled Optimization at Scale (TILOS), NSF award CCF-2112665.

\end{document}